\documentclass[10pt,twocolumn,letterpaper]{article}
\usepackage[pagenumbers]{iccv} 

\definecolor{iccvblue}{rgb}{0.21,0.49,0.74}
\usepackage[pagebackref,breaklinks,colorlinks,allcolors=iccvblue]{hyperref}
\usepackage{multirow}
\usepackage{float}
\usepackage{makecell}
\usepackage{tabularx}
\newcolumntype{Y}{>{\raggedright\arraybackslash}X}

\def\paperID{3862} 
\def\confName{ICCV}
\def\confYear{2025}

\title{LaRender: Training-Free Occlusion Control in Image Generation via Latent Rendering}

\author{Xiaohang Zhan\\
Tencent\\
{\tt\small xiaohangzhan@outlook.com}
\and
Dingming Liu\\
Tencent\\
{\tt\small dingmingliu3722@gmail.com}
}

\begin{document}
\twocolumn[{%
\renewcommand\twocolumn[1][]{#1}%
\maketitle
\begin{center}
    \centering
    \captionsetup{type=figure}
    \vspace{-5pt}
    \includegraphics[width=1.0\textwidth]{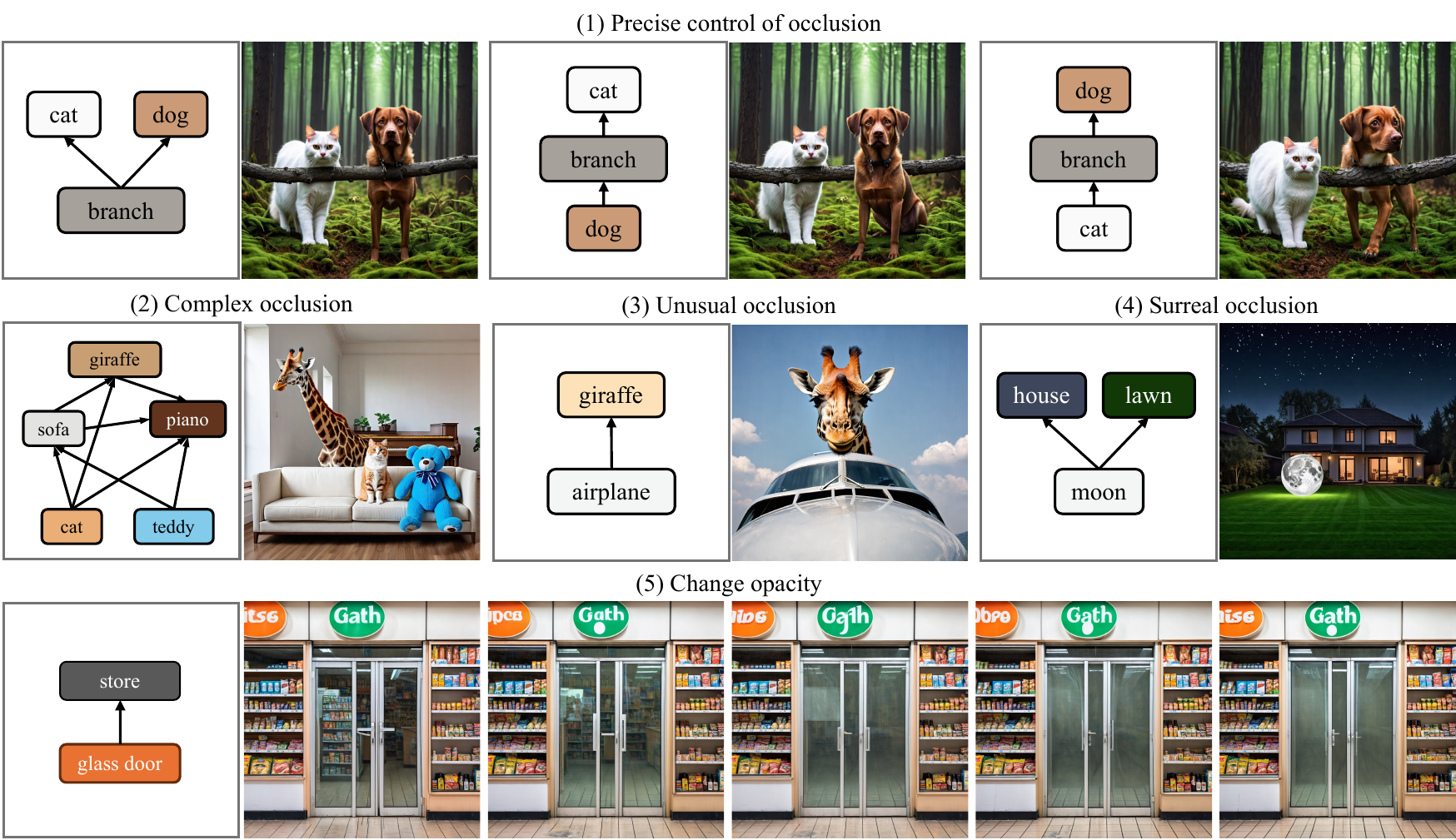}
    \captionof{figure}{
    This work enables precise control of occlusion relationships among objects, and performs well in situations including complex, unusual and surreal occlusion. It also enables rich effects such as opacity control. All of these do not require training or fine-tuning. The background is omitted from the graph for clarity. Project page: \url{https://xiaohangzhan.github.io/projects/larender/}.
    \label{fig:teaser}}
\end{center}%
}]

\begin{abstract}

We propose a novel training-free image generation algorithm that precisely controls the occlusion relationships between objects in an image. Existing image generation methods typically rely on prompts to influence occlusion, which often lack precision. While layout-to-image methods provide control over object locations, they fail to address occlusion relationships explicitly. Given a pre-trained image diffusion model, our method leverages volume rendering principles to ``render'' the scene in latent space, guided by occlusion relationships and the estimated transmittance of objects. This approach does not require retraining or fine-tuning the image diffusion model, yet it enables accurate occlusion control due to its physics-grounded foundation. In extensive experiments, our method significantly outperforms existing approaches in terms of occlusion accuracy. Furthermore, we demonstrate that by adjusting the opacities of objects or concepts during rendering, our method can achieve a variety of effects, such as altering the transparency of objects, the density of mass (\textit{e.g.,} forests), the concentration of particles (\textit{e.g.,} rain, fog), the intensity of light, and the strength of lens effects, \textit{etc}.

\end{abstract}    
\section{Introduction}

Occlusion control is a critical aspect of image generation, particularly in applications such as advertising content creation, concept design, and complex scene generation, where the spatial arrangement and interaction of objects must be accurately represented. Precise occlusion control ensures that objects are correctly layered and interact in a visually coherent manner, which is essential for creating realistic and immersive scenes.

Despite the advancements in image generation techniques, current state-of-the-art methods struggle to provide precise occlusion control. Existing text-to-image approaches~\cite{rombach2022high,saharia2022photorealistic,peebles2023scalable,chen2023pixart,podell2023sdxl,nichol2021glide,ramesh2021zero,ramesh2022hierarchical} rely on text prompts to control occlusion, such as \textit{``a dog behind a cat, and a bird is in front of the cat''}. However, in practice, surprisingly, even the state-of-the-art method~\cite{flux2024} performs poorly in occlusion control, especially in complex scenes with multiple objects occluding each other.

The other related branch is layout-to-image generation~\cite{zhou20243dis,zhou2024migc,chen2024training,xie2023boxdiff,zheng2023layoutdiffusion,bar2023multidiffusion,li2023gligen}. Given the layout that is usually represented as bounding boxes, these methods can generate complex scenes. Since layout shares some common priors with occlusion, layout-to-image models have the potential to generate images with reasonable occlusion relationships. However, due to the complexity of the occlusion phenomenon, these methods cannot accurately control the occlusion relationships. As shown in Figure~\ref{fig:teaser} (1), each object shares the same bounding box in different occlusion cases.
Another trivial solution, though unexplored in existing work, could involve training or fine-tuning an image generation model conditioned on occlusion relationships. However, this approach requires additional paired data of images and ground-truth occlusions, which are often expensive to collect.
 
The occlusion phenomenon shares the same essence as 3D rendering, whether from the perspective of human vision or a camera. This insight leads us to explore the relationship between 3D rendering and occlusion control in image generation.
By leveraging principles of Volumetric Rendering, we propose \textbf{LaRender}, a non-parametric training-free method that effectively addresses the occlusion control problem in image generation.
As shown in Figure ~\ref{fig:teaser}, LaRender achieves precise occlusion control without the need for retraining or fine-tuning the image generation model. The method performs well even in complex, unusual, or surreal occlusion scenarios. Inheriting its essence from rendering, LaRender also enables control over object opacity, producing rich effects such as changing the transparency of objects, the density of mass (\textit{e.g.,} forests), the concentration of particles (\textit{e.g.,} rain, fog), the intensity of light, and the strength of lens effects. Please find more results in Figure~\ref{fig:density_control}. These capabilities come as a``free lunch'', requiring no additional data or training.

In brief, we propose a non-parametric Latent Rendering mechanism that replaces vanilla cross-attention layers of a pre-trained image diffusion model. Given an occlusion graph, our method first sorts objects from bottom to top and rearranges the objects' latent features, \textit{i.e.,}, the hidden states from the denosing network, followed by a virtual camera facing the latent features. We then perform Latent Rendering, which borrows principles from volumetric rendering but operates at the latent level rather than the pixel level. In this way, latent features of objects are integrated according to physical rules that consider both occlusion relationships and object transmittance.

The key contributions of our work are as follows:
\begin{itemize}
    \item Inspired by the shared essence of occlusion and 3D rendering, we propose a novel mechanism, Latent Rendering, which for the first time, addresses the occlusion control problem in image generation without the need for training.
    \item Beyond precise occlusion control, we observe high-quality, physics-grounded visual effects enabled by LaRender, which potentially inspire new applications.
\end{itemize}
\section{Related Work}

\subsection{Image Generation}
\noindent\textbf{Text-to-Image Generation.}
Recent advances in diffusion models~\cite{rombach2022high,saharia2022photorealistic,peebles2023scalable,chen2023pixart,podell2023sdxl,nichol2021glide,ramesh2021zero,ramesh2022hierarchical} have significantly improved text-to-image generation, enabling diverse, high-quality, and controllable outputs. Open-source projects like Stable Diffusion~\cite{rombach2022high}, Stable Diffusion XL~\cite{podell2023sdxl}, and FLUX~\cite{flux2024} have further facilitated research in this area.

\noindent\textbf{Layout-to-image Generation.}
Layout-to-image methods~\cite{zhou20243dis,zhou2024migc,chen2024training,xie2023boxdiff,zheng2023layoutdiffusion,bar2023multidiffusion,li2023gligen,wang2024spotactor} control object spatial layouts to generate complex scenes. Some~\cite{li2023gligen,zhou2024migc,zhou20243dis} fine-tune diffusion models using datasets with bounding box annotations, such as MS COCO~\cite{lin2014microsoft}. Specifically, 3DIS~\cite{zhou20243dis} generates depth maps first, then renders textures from depth. However, depth does not equate to occlusion~\cite{zhan2020self}, and object depths remain uncontrollable. Others~\cite{bar2023multidiffusion,xie2023boxdiff,chen2024training} proposed training-free methods focusing on object positioning. While our method uses bounding boxes to compute transmittance, it prioritizes occlusion control over precise positioning.

\noindent\textbf{Scene graph-to-image Generation.}
Scene graphs represent objects and relationships. There are also low-level relationships, such as ``in front of, behind'' that share some common concepts with occlusion. However, these concepts focus on relative spatial positioning rather than the occlusion from the photographic perspective. For example, ``a boy standing in front of a girl'' can be a scene where they stand face to face, one on the left and the other on the right, with no visual occlusion. Moreover, scene graph-to-image generation methods~\cite{johnson2018image,tripathi2019using,yang2022diffusionsg} rely on datasets that have scene graph annotations, limiting their applicability.

\noindent\textbf{Multi-layer Image Generation.}
MULAN~\cite{tudosiu2024mulan} introduces a multi-layer dataset with RGBA decomposed objects. Zhang \textit{et al.}~\cite{zhang2024transparent} fine-tune latent diffusion models on 1M transparent image layer pairs for transparent image generation. LayerFusion~\cite{dalva2024layerfusion} generates foreground, background and blended images from text prompts. However, these methods are not designed for complex scenes with multiple occluding objects.

\noindent\textbf{3D-Aware Image Generation.}
Previous works~\cite{niemeyer2021giraffe,schwarz2020graf,chan2021pi,gu2021stylenerf,chan2022efficient} integrate GANs with neural radiance fields for 3D-aware image generation. They transplant learnable volume or triplane representations in GANs,  allowing the synthesis of multi-view-consistent images. However, they are typically limited to constrained domains such as faces, cars, or animals. Recently, ViewDiff~\cite{hollein2024viewdiff} leverages multi-view supervision and volume rendering to generate 3D-consistent images in more diverse settings. In contrast, our approach introduces a non-parametric volumetric rendering mechanism applied in the latent space of diffusion models, enabling generating complex scenes in a training-free manner.

\subsection{Occlusion Handling}
Occlusion handling is a classic challenge in computer vision, particularly in complex scene understanding. Datasets with occlusion annotations~\cite{zhu2017semantic,qi2019amodal} and self-supervised methods~\cite{zhan2020self} have laid the foundation of occlusion handling from the perspective of perception. Follow up works~\cite{ke2021deep,zhou2021human,zhan2024amodal,liu2024object,ling2020variational,zhan2022tri} further improve de-occlusion. However, from the generative perspective, occlusion handling remains a challenge. We are the first to address this challenge in the generative domain.
\section{Methodology}


\subsection{Preliminaries of Volume Rendering}
\label{sec:vr}
Volume rendering~\cite{drebin1988volume} computes the accumulated color of a pixel via integral of the volume color density along the camera ray:
\begin{equation}
\mathbf{C} = \int_{t_n}^{t_f} T(t) \sigma(t) \mathbf{c}(t) dt, \text{where } T(t)=\exp(-\int_{t_n}^{t}\sigma(s)ds),
\end{equation}
where $t$ is the position along the camera ray with near bounds $t_n$ and far bounds $t_f$, $T(t)$ denotes the accumulated transmittance along the ray from $t_n$ to $t$, $sigma(t)$ is the volume density, $\mathbf{c}(t)$ is the color of position $t$.

NeRF~\cite{mildenhall2021nerf} numerically estimates this continuous integral via quadrature, resulting in the formula as:

\begin{equation}
\mathbf{\hat{C}} = \sum_{i=1}^{N} T_i (1 - \exp(-\sigma_i\delta_i)) \mathbf{c}_i,  \text{where } T_i=\exp(-\sum_{j=1}^{i-1}\sigma_j\delta_j),
\end{equation}
where $i=1,2,...,N$ is the sampled positions in $[t_n, t_f]$, $\delta_i=t_{i+1}-t_i$ is the distance between the adjacent samples.

\subsection{Latent Rendering for Occlusion Control}
\label{sec:lr}
\begin{figure*}[t]
  \centering
  \includegraphics[width=\textwidth]{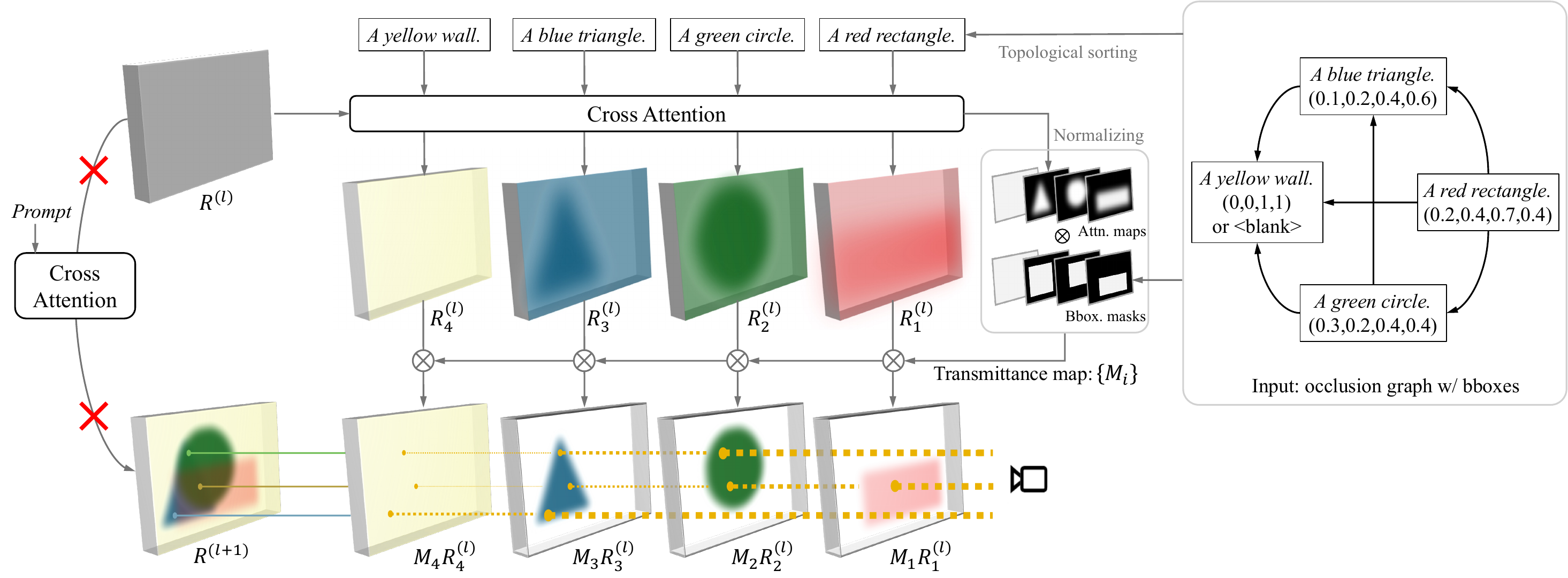}
  \caption{\textbf{LaRender} simply replaces vanilla cross attention layers to Latent Rendering layers. Given an occlusion graph with bounding boxes provided by users or parsed by LLMs, we rearrange these objects from back to front via topological sorting. For each cross-attention layer in a trained image diffusion model, we allow the input feature attend to each object prompt, obtaining object-wise latent features (hidden states). Then we estimate the transmittance map $\mathbf{M}_i$ for each object from its attention map and bounding box. Subsequently, we position an orthographic camera above the top object, facing the latent features. Finally, we apply the Latent Rendering formula to obtain the output scene representation that physically ensures occlusion relationships. The width of the orange dashed lines depicts the change of the accumulated transmittance $\mathbf{T}_i$ (initially to be 1) in different positions, and the colored solid lines represent the integrated latent features through their respective camera rays.}
  \label{fig:framework}
\end{figure*}

The key idea of Latent Rendering is to adapt the volume rendering formula to ``render'' a stack of layered latent features of objects, resulting in a fusion of object latent features that physically ensures occlusion relationships.

\noindent\textbf{Latent re-arrangement}. As shown in Figure~\ref{fig:framework}, given the occlusion graph and the bounding boxes of each object, we first apply topological sorting to sort objects from top to bottom, numbered as $1,2,...,N$. Then given a trained image diffusion model, for the $l$-th cross-attention layer, we modify it to allow the latent representation $\mathbf{R}^{(l)}$ to attend to the prompt of each object, resulting in $N$ object-wise latent features $\mathbf{R}_i^{(l)}$.
Next, we draw an analogy to Volume Rendering: we stack all $\mathbf{R}_i^{(l)}$ from back to front, and position a virtual orthographic camera above the top object, facing the latent features. Since the camera is orthographic, the distance between adjacent latent features is not necessarily defined - we omitted the distance $\delta_i$ in our rendering formula. We use $\mathbf{R}_i^{(l)}$ to replace the color of all grid sampling positions in Volume Rendering, and define semantic density $\sigma_i$ (a scalar) to replace the volume density in Volume Rendering.

\noindent\textbf{Transmittance map.} While topological sorting reflects the occlusion order, the specific regions of occlusion are unknown. There regions are essentially determined by the transmittance map - the spatial distribution of transmittance for each layer of objects. 
To estimate the transmittance map, we primarily use bounding box masks provided by users or LLMs. However, transmittance can be inaccurate in the interval between the bounding box mask and the actual area occupied by the object. To address this, we reuse the cross-attention maps from the aforementioned cross-attention module. First, we extract the index of the subject token via Dependency Parsing~\cite{honnibal2017spacy}, and normalize the cross-attention map of the subject token to rescale its minimum and maximum to 0 and 1. Then, we perform an element-wise multiplication of this normalized map with the bounding box mask, yielding the final transmittance map.

\noindent\textbf{Latent Rendering.} Afterwards, we formally define Latent Rendering, as follows:
\begin{equation}
\begin{aligned}
\mathbf{R}^{(l+1)} &=\frac{1}{\mathbf{S}} \sum_{i=1}^{N} \mathbf{T}_i (1 - \exp(-\sigma_i)) \mathbf{M}_i \mathbf{R}_i^{(l)}, \\
\mathbf{S} &= \sum_{i=1}^{N} \mathbf{T}_i (1 - \exp(-\sigma_i)) \mathbf{M}_i,\\
\mathbf{T}_i &= \exp\left(-\sum_{j=1}^{i-1} \mathbf{M}_j \sigma_j\right),
\end{aligned}
\end{equation}
where $\mathbf{R}^{(l+1)}$ is the updated latent features to be fed into the next layer, $\mathbf{S}$ is the normalization term to prevent the output from deviating from the original latent distribution, $\mathbf{M}_i$ is the transmittance map, $\mathbf{T}_i$ is the accumulated transmittance map that describes the visibility of all pixels of object $i$ from the virtual camera. Scalar $\sigma_i >0$ is the semantic density of object $i$, similar to the volume density in volume rendering, we will discuss it later.

For methodological simplicity, we replace all cross-attention layers in the denoising network with our proposed Latent Rendering and follow the standard sampling steps to generate the image.
Though we observed different behaviors of Latent Rendering in varying layers, we left it for future researchers to explore.
Notably, Latent Rendering does not introduce any learnable parameters, making the framework entirely training-free. In this way, we generate images in a physically grounded manner that faithfully reflects occlusion relationships.




\noindent\textbf{Density Scheduling.} 
The semantic density $\sigma_i$, different from opacity, reflects the semantic strength of object concept $i$.
We observed that in early denoising steps when the latent $\mathbf{R}^{l}_i$ does not yet have a strong concept of the object, Latent Rendering can disrupt this fragile concept by mixing the latent features.
We also found that if $\sigma_i \rightarrow +\infty$, Latent Rendering degrades to ``opaque mode'' where all latent features become non-transparent, preventing them from mixing with each other. The mode is particularly suitable for early denoising steps. As revealed in ~\cite{balaji2022ediff}, object concepts rapidly form during denosing, and late steps focus more on refining image quality rather than establishing concepts. Therefore, during middle and late steps, Latent Rendering can effectively distinguish different concepts even after mixing, and ``opaque mode'' is no longer required.
Therefore, we introduce a fast-to-slow descent schedule to control the semantic density. Specifically, we define $\sigma_i$ as a simple inverse proportional function of the diffusion step $t$:
\begin{equation}
\label{eq:schedule}
\sigma_i(t) =  \frac{D_i T}{T + 1 - t},
\end{equation}
where the scalar $D_i\geq 0$ is the semantic density of object $i$ to be determined by users, and $T$ is the total number of diffusion steps, $t=T,T-1, ..., 1$ is the current denoising step. During this schedule, initially $\sigma_i(T)=D_iT$ is a sufficiently large value, making Latent Rendering approximate an ``opaque mode''. At the last step, $\sigma_i(1)=D_i$, which means Latent Rendering converges to the target density $D_i$. We compare different schedules in the experiments.

\noindent\textbf{Input mode.} In cases where it is difficult for users to specify the occlusion graph and bounding boxes, we provide an alternative option: use an LLM to parse the original prompt to output the occlusion graph and the bounding boxes. We tried several LLMs, including those from GPT series, DeepSeek, and Claude, and observed over 95\% accuracies in parsing the ordering, along with reasonable recommendation of bounding boxes. Thus, users can either simply write initial prompts or edit the occlusion and object positions after the LLM's parsing. Besides, if the users do not wish to specify the background such as a \textit{``wall''} in the case of Figure~\ref{fig:framework}, they can use a blank prompt \textit{``''}.



\subsection{Control Occlusion and Semantic Opacity}
\label{sec:ctrl}
Via modifying the bounding boxes and the occlusion graph, we can easily control the precise occlusion relationships among objects, as well as rough positions. Note that though LaRender is able to control rough positions, it does not focus on improving the layout control accuracy - it is orthogonal to layout control methods.

Since the semantic density $D_i$ is an unbounded positive value, we let $\alpha_i = 1 - \exp(-D_i)$ as the semantic opacity, and control $\alpha_i\in[0,1)$ to observe the effects of semantic density. In extreme cases, when $\alpha_i = 0$, the object is entirely transparent and will disappear from the generated image; when $\alpha_i \rightarrow 1$, then $D_i\rightarrow +\infty$, that means the object is opaque with high density, and no pattern of other objects can appear through it.

When $\alpha_i\in(0,1)$, does it mean the object is semi-transparent? Not exactly. Since we perform rendering in latent level rather than in RGB space, the semantic density should demonstrate high-level ``semi-transparency''. In fact, during our experiments, we observe interesting high-level ``semi-transparency'' phenomena, including changing density of mass (\textit{e.g.}, forests), concentration of particles (\textit{e.g.}, rain, fog), intensity of light, strength of lens effects, \textit{etc}. Of course, if the semantic density of an object is strongly correlated with its transparency, then the physical transparency can naturally be controlled as well.
\section{Experiments}
\subsection{Implementation Details}

We used Stable Diffusion XL (SDXL)~\cite{podell2023sdxl} pre-trained by IterComp~\cite{zhang2024itercomp} as the base model for its balance of quality and efficiency, though our method is compatible with varying diffusion-based image generation network. Results based on FLUX.1-dev~\cite{flux2024} are included in the supplementary materials. To ensure fairness and avoid evaluation distractions, we did not employ prompt modification tricks or negative prompts. The only hyperparameter is the semantic opacity $\alpha$ for each object. We set $\alpha=0.8$ for all quantitative comparisons and varied it from 0.1 to 0.9 in semantic density control experiments (Figure~\ref{fig:density_control}). All other inference hyperparameters, including the denoising step T in Equation~\ref{eq:schedule}, followed the base model’s default settings. For occlusion graph and bounding box generation, we used DeepSeek R1~\cite{guo2025deepseek}.

\begin{table*}[ht]
\setlength{\tabcolsep}{3.5pt}
\centering
\caption{Quantitative comparisons on T2I-CompBench++ (3D Spatial) dataset and our proposed RealOcc dataset. Though there is no existing work on occlusion control, we maximized the ability of occlusion control of state-of-the-art text-to-image (SDXL, FLUX) and state-of-the-art layout-to-image (MIGC, 3DIS) models via prompt and layout design. Our method outperforms all baselines in occlusion related indicators,\textit{i.e.}, ``UniDet'' and ``User study'', and shows minimal degradation in CLIP score. The denoising step is 25 for LaRender and SDXL, and 20 for FLUX and 3DIS. AUR and HPSR are calculated with 95\% t-distribution Confidence Intervals.}
\scalebox{0.9}{
\begin{tabular}{l|l|cccc|ccc|c}
\hline
\multirow{3}{*}{Methods}& \multirow{3}{*}{Control}&\multicolumn{4}{c|}{T2I-CompBench++ - 3D Spatial(val)}   & \multicolumn{3}{c|}{RealOcc} &\multirow{3}{*}{time (s)} \\\cline{3-9}
&& \multirow{2}{*}{UniDet$\uparrow$} & \multicolumn{2}{c}{User study} & \multirow{2}{*}{CLIP score$\uparrow$} & \multicolumn{2}{c}{User study} & \multirow{2}{*}{CLIP score$\uparrow$} &  \\
&       &  & AUR$\uparrow$    & HPSR$\uparrow$   &    & AUR$\uparrow$   & HPSR$\uparrow$   &    & \multicolumn{1}{c}{}      \\\hline
SDXL~\cite{podell2023sdxl} & text& 0.357 & 2.36 ± 0.49 & 0.322 ± 0.051 & \textbf{31.12} & 3.08 ± 0.40 & 0.396 ± 0.045  &  \textbf{29.42}  & 7.44 \\
FLUX~\cite{flux2024} & text & 0.401 & 1.94 ± 0.42          & 0.293 ± 0.023  & 30.86 & 1.36 ± 0.25 & 0.244 ± 0.051  &  28.13  & 122 \\\hline
MIGC~\cite{zhou2024migc} & layout & 0.373 & 3.56 ± 0.52          & 0.448 ± 0.068  & 30.73 & 3.31 ± 0.51 & 0.433 ± 0.070  &28.49  & 11.1 \\
3DIS~\cite{zhou20243dis} & layout & 0.337& 2.19 ± 0.33          & 0.311 ± 0.051 & 30.11 & 2.44 ± 0.49 & 0.344 ± 0.058 &28.07& 104 \\\hline
LaRender (ours)   & occlusion  & \textbf{0.416}& \textbf{4.94} ± 0.11          & \textbf{0.767} ± 0.070 & 30.98& \textbf{4.81} ± 0.29 & \textbf{0.767} ± 0.094 & 28.30 & 7.46 \\\hline
\end{tabular}
}
\label{tab:compare}
\end{table*}


\begin{figure*}[!h]
\centering
\begin{minipage}{0.96\columnwidth}
\centering
\captionof{table}{The impact of different density schedules.}
\vspace{-5pt}
\small
\begin{tabular}{l|l|cc}
    \hline
        Schedule  & Formula & UniDet$\uparrow$& CLIP score$\uparrow$\\\hline
        Fixed opaque mode & $\sigma_i(t)=D_iT$ & 0.240 & 28.32 \\
        Fixed density & $\sigma_i(t)=D_i$       & 0.393 & 30.44 \\\hline
        \makecell{Inverse proportional\\function (adopted) } & $\sigma_i(t) =\frac{D_i T}{T + 1 - t}$ & \textbf{0.416} & \textbf{30.98} \\\hline
    \end{tabular}
\label{tab:schedule}
\end{minipage}
\hfill
\begin{minipage}{0.96\columnwidth}
\centering
\captionof{table}{The impact of the attention maps in computing transmittance maps.}
\begin{tabular}{l|cc}
    \hline
        Choice  & UniDet$\uparrow$& CLIP score$\uparrow$\\\hline
        LaRender w/o attn. map &0.395 & \textbf{31.00} \\
        LaRender w/ attn. map &\textbf{ 0.416} & 30.98 \\\hline
    \end{tabular}
    \label{tab:attn_map}
\end{minipage}
\end{figure*}



\begin{figure*}[!h]
  \centering
  \includegraphics[width=0.85\textwidth]{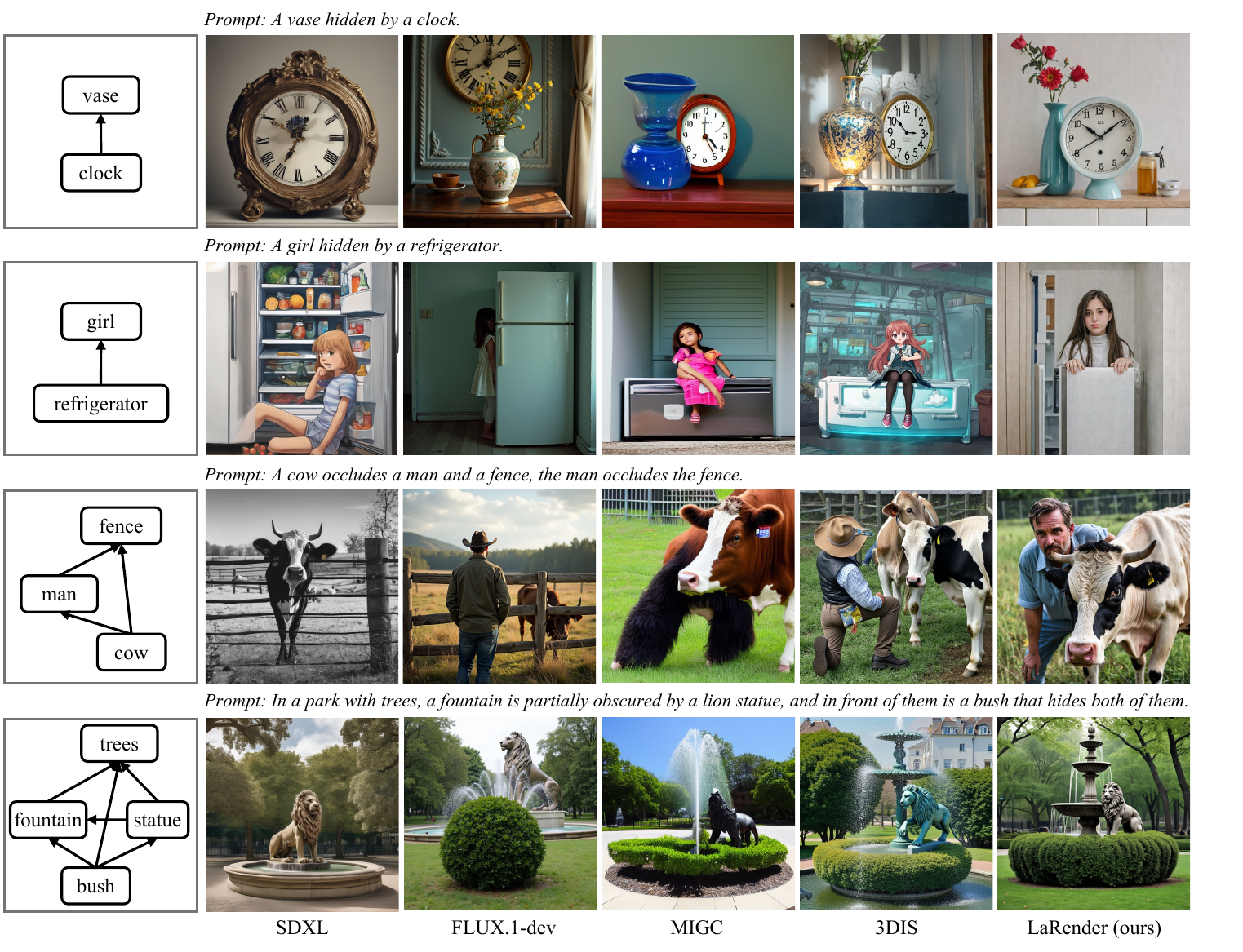}
  \caption{Qualitative comparison. The first column represents the occlusion graph. The cases in top two rows are results from T2I-CompBench++ (3D Spatial) dataset, the third row is a case in the RealOcc dataset, the case in last row is invented by ourselves.}
  \label{fig:compare}
\end{figure*}

\subsection{Evaluation Protocols}

\noindent\textbf{Baselines.}
There is no existing work that aims at occlusion control in image generation. However, text-to-image methods should be able to control occlusion via prompting, and layout-to-image models may generate images adhering to plausible occlusion relationships via carefully designed layouts, especially 3DIS~\cite{zhou20243dis} that uses depth as an intermediate signal. We maximized their possibility of occlusion control via prompt and layout design. The baselines we used are as follows:
\begin{enumerate}
    \item SDXL~\cite{podell2023sdxl}. A widely used text-to-image model.
    \item FLUX.1-dev~\cite{flux2024}. A state-of-the-art open-source text-to-image model.
    \item MIGC~\cite{zhou2024migc}. A state-of-the-art layout-to-image model fine-tuned on the COCO dataset.
    \item 3DIS~\cite{zhou20243dis}. A state-of-the-art layout-to-depth-to-image model fine-tuned on COCO. We used the FLUX version of the depth-to-image stage for the maximal quality.
\end{enumerate}

\noindent\textbf{Evaluation data and metrics.}
We used the data below:
\begin{enumerate}
    \item \textbf{T2I-CompBench++}~\cite{huang2025t2i} - 3D spatial relationship evaluation. The validation split contains 300 prompts, each describing two objects with simple spatial relationships (\textit{e.g.}, "in front of", "behind", "hidden by").  We used DeepSeek to generate bounding boxes and manually adjusted failure cases for methods requiring box inputs. We followed the standard UniDet-based metric, which combines object detection and depth estimation to infer object order.
    \item \textbf{RealOcc.} To complement T2I-CompBench++, which lacks real layouts and only includes two-object prompts, we created RealOcc, a dataset with real-world bounding boxes. We curated images with 2-5 objects from the COCOA~\cite{zhu2017semantic} validation set, filtering out extremely small or large bounding boxes and invalid annotations such as ``background'', ensuring each object has at least one occlusion relationship inferred from COCOA's occlusion annotations, and ensuring a balanced distribution of number of objects. This results in 70 examples with 249 objects with amodal bounding boxes and 261 occlusion pairs. While small, it serves as a valuable complement to T2I-CompBench++. Detailed statistics are provided in the supplementary materials.
\end{enumerate}

\begin{figure*}[ht]
  \centering
  \includegraphics[width=0.9\textwidth]{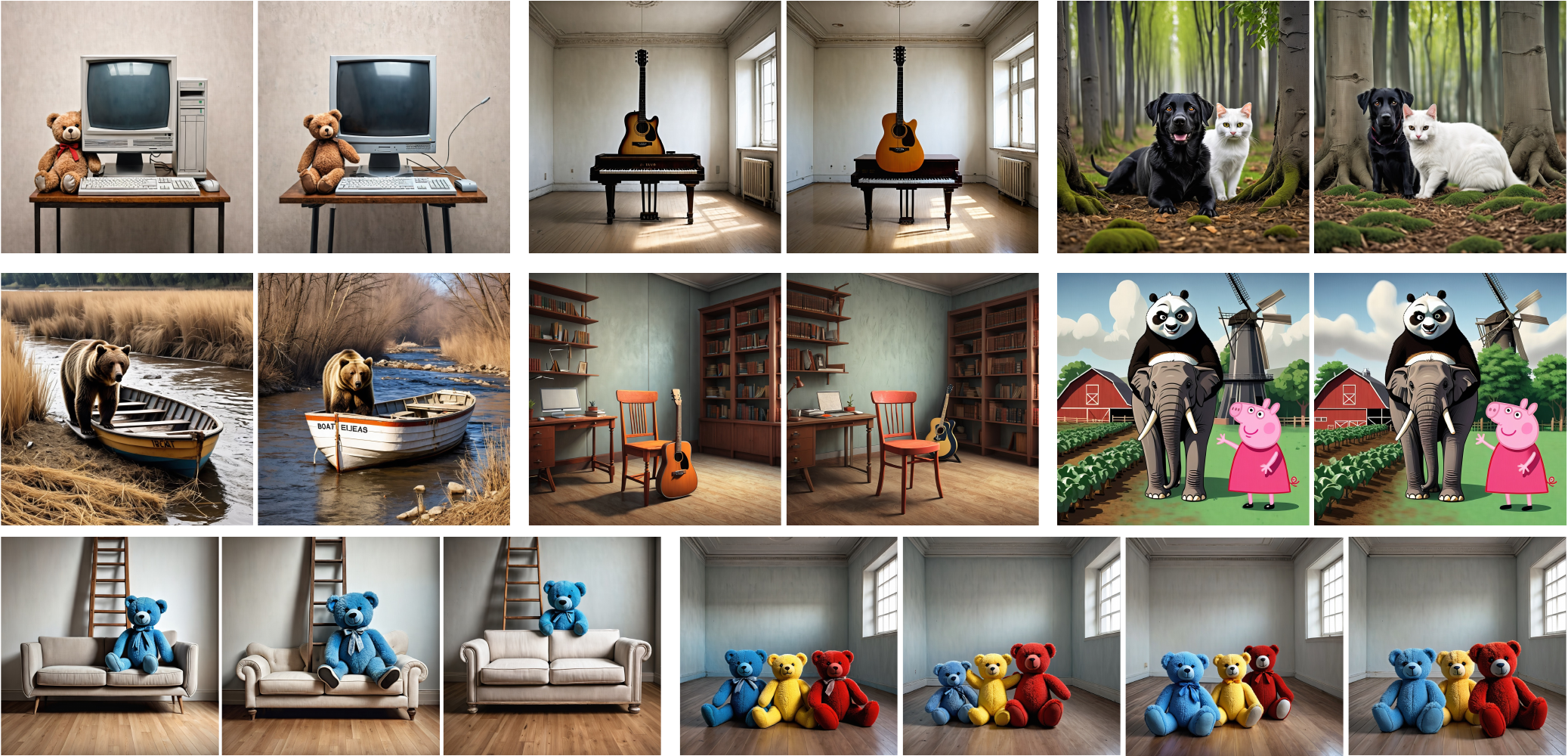}
  \caption{Our generated results given similar layouts but different occlusion relationships.}
  \label{fig:same_layout}
\end{figure*}

Additionally, we conducted user studies as a more comprehensive evaluation of occlusion relationships. We designed 15 comparison samples per dataset and distributed over 30 questionnaires. Following ControlNet~\cite{zhang2023adding}, we reported Average User Ranking (AUR) and Human-Perceived Success Rate (HPSR) in Table~\ref{tab:compare}. To ensure Latent Rendering does not degrade the base model's generative ability, we also evaluated the CLIP score for both datasets. We did not compute FID, since the former dataset has no real images and the latter contains too few images that are unable to fulfill FID's minimal requirement of 3K images - it is hard to scale up this dataset due to the lack of amodal annotations.

\subsection{Results and Comparisons}

\noindent\textbf{Quantitative comparison.} As shown in Table~\ref{tab:compare}, we obtained results of baseline methods on T2I-CompBench++ (3D spatial) and RealOcc datasets with their official code. 
Compared with state-of-the-art text-to-image and layout-to-image methods, LaRender achieves the best performance on occlusion related indicators, \textit{i.e.}, UniDet and User study.
Besides, we observe a slight decrease on CLIP score compared with SDXL. It is mainly because LaRender receives a list of objects as input, instead of a complete text prompt.

\noindent\textbf{Qualitative comparison}. As shown in Figure~\ref{fig:compare}, for the cases with different numbers of objects, our method consistently produces precise occlusion controlled results. Though instructed by descriptions about occlusion, Text-to-image methods SDXL and FLUX cannot generate images with correct occlusion, especially when the number of objects is larger than two. Layout-to-image methods MIGC and 3DIS, though generate images with correct positions, cannot deal with occlusion relationships.

\subsection{Analysis}

\noindent\textbf{Ablation study.} We studied the effectiveness of density scheduling in Latent Rendering. As shown in Figure~\ref{tab:schedule}, in fixed opaque mode when $\sigma_i(t)$ is always a sufficiently large value, LaRender is degraded and the performance is poor. If we fix the density as a normal value during denoising, as discussed in the methodology section, the concepts can be destroyed, thus we observed decrease of the performance. Our adopted inverse proportional function performs the best on both UniDet and CLIP score. We also studied the impact of the cross-attention maps in computing the transmittance maps. As shown in Table~\ref{tab:attn_map}, with attention map to shape the contour of objects in latent space, the transmittance map is more accurate, thus we observed higher performance in occlusion control, while bringing almost no impact on CLIP score. See supplementary materials for visualized results.

\noindent\textbf{Time consumption.} Compared with the base model SDXL, LaRender's increase of inference time is negligible. This is because the computation of LaRender involves element-wise multiplication and summation, which can be fully parallelized.

\begin{figure*}[ht]
  \centering
  \includegraphics[width=0.9\textwidth]{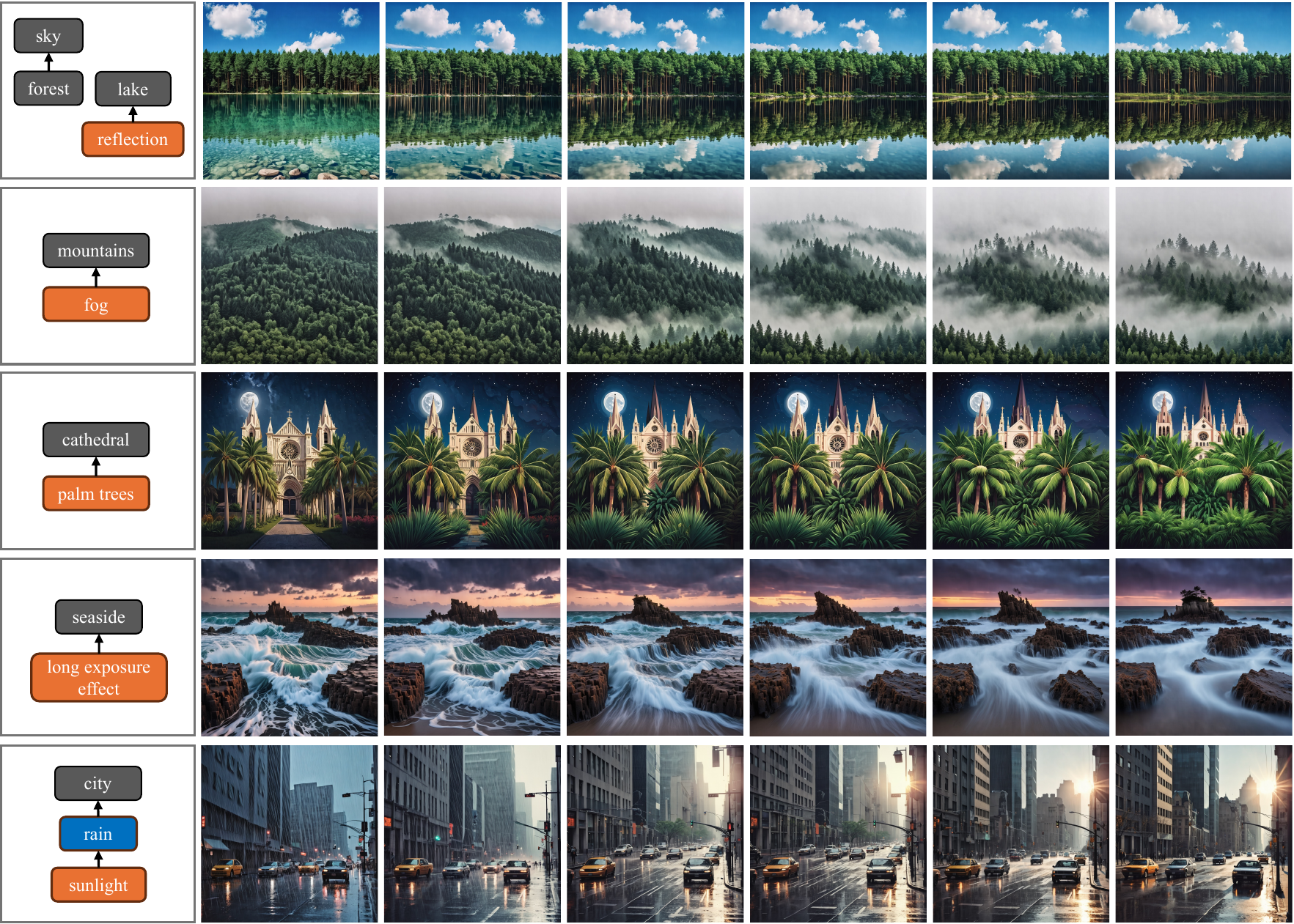}
  \vspace{-10pt}
  \caption{Controlling the concept density/strength via adjusting the semantic opacity $\alpha$. In the first column, gray concepts keep the same opacities, orange concepts are increasing their opacities, and the blue concept is decreasing its opacity. Please find the dual-element grid figure for the last case in the supplementary materials.}
  \label{fig:density_control}
\end{figure*}

\begin{figure*}
    \centering
  \includegraphics[width=0.9\textwidth]{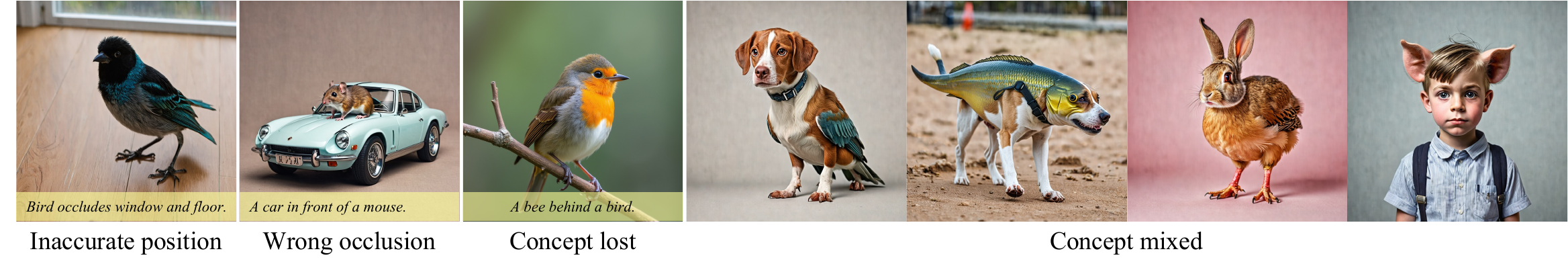}
  \vspace{-10pt}
  \caption{Failure cases of LaRender. See Section~\ref{sec:conclusion} for analysis.}
  \label{fig:failure}
\end{figure*}

\subsection{Applications}

\noindent\textbf{Controlling occlusion}.
With LaRender, we are able to freely control the occlusion relationships among multiple objects, even in similar layouts. As shown in Figure~\ref{fig:same_layout}, given similar layouts but different occlusion relationships, LaRender consistently generates high-quality and occlusion-precise images. Please find more results including LaRender with FLUX in the supplementary materials.

\noindent\textbf{Interactions}.
Normally LaRender accepts independent object prompts, we find that it supports interactions by assigning a full prompt to the background. The visual results are shown in the supplementary materials.

\noindent\textbf{Adjusting Semantic Opacity}.
In addition to controlling occlusion relationships, LaRender exposes an adjustable parameter, the semantic opacity. By controlling the semantic opacities, we observed interesting ``semi-transparency'' phenomena. As shown in 
Figure~\ref{fig:density_control}, the semantic opacity affects the appearance of different objects in different ways, including changing the transparency of glass doors (Figure~\ref{fig:teaser}), the concentration of fog, the density of palm trees. Surprisingly, it can even be applied to non-object concepts, such as changing the strength of reflection, long-exposure effect and sunlight.
\section{Conclusion, Limitations and Social Impact}
\label{sec:conclusion}
In \textbf{conclusion}, we proposed a novel non-parametric mechanism, Latent Rendering, \textit{abbr.} LaRender, that is able to provide precise control of occlusion relationships among objects in image generation in a training-free way. Extensive experiments prove its effectiveness and efficiency. We also observed interesting concept strength controlling effects brought by LaRender.

\noindent\textbf{Limitations}. The occlusion results can be wrong when the layout is not reasonable. The bounding boxes are rough hints of positioning, we are not pursuing accurate bounding box control in this paper. Additionally, sometimes the concepts can be lost or mixed, as shown in Figure~\ref{fig:failure}. That is because the latent features can happen to be mixed or erased rather than partially occluded to satisfy its generative prior. Similar issues have been discussed in~\cite{wei2024enhancing}.

\noindent\textbf{Social Impact}. The proposed method, as an image generation approach, could potentially be misused to create deceptive or misleading visual content, raising concerns about privacy, security, and misinformation. To mitigate these risks, we recommend implementing strict usage guidelines, monitoring for misuse, and developing robust detection mechanisms for identifying generated images.
\clearpage
{
    \small
    \bibliographystyle{ieeenat_fullname}
    \bibliography{main}
}

\clearpage
\appendix
\section{The RealOCC Dataset}

\begin{figure}[!h]
    \centering
    \includegraphics[width=\linewidth]{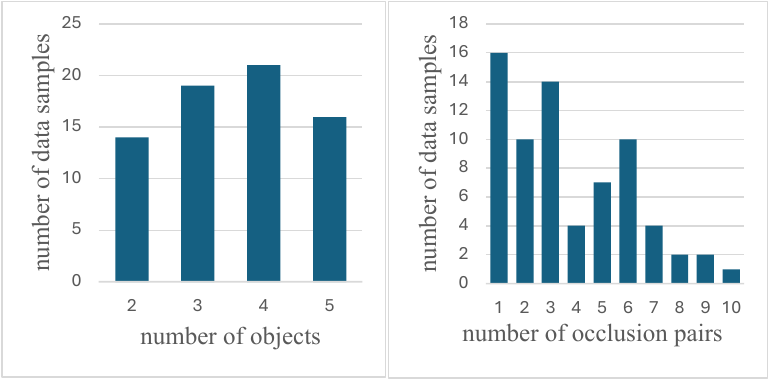}
    \caption{The statistics of the RealOcc dataset.}
    \label{fig:statistics}
\end{figure}

\noindent\textbf{Statistics}. As shown in Figure~\ref{fig:statistics}, the RealOCC dataset maintains a balanced distribution of images with 2 to 5 objects, capturing diverse occlusion scenarios. Images with more objects usually contain fewer occlusion pairs, which reflects real-world patterns. On average, each image includes 3.56 objects and 3.73 occlusion pairs, offering varied and realistic examples for evaluating occlusion relationships. This distribution ensures the dataset is both diverse and practical for testing occlusion handling methods.

\begin{figure}[t]
  \centering
  \includegraphics[width=\linewidth]{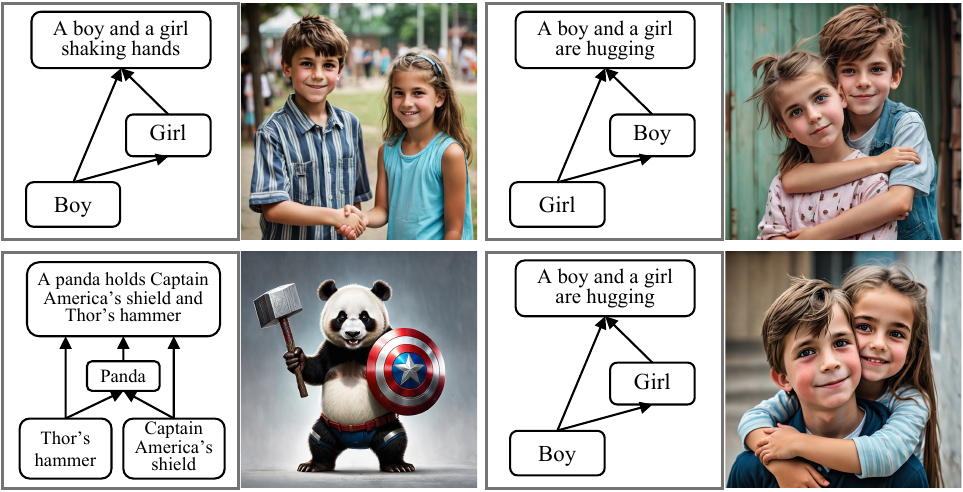}
   \caption{Examples of generating inter-object interactions via background prompts.}
   \label{fig:interactions}
\end{figure}

\begin{figure*}[t]
    \centering
    \includegraphics[width=\linewidth]{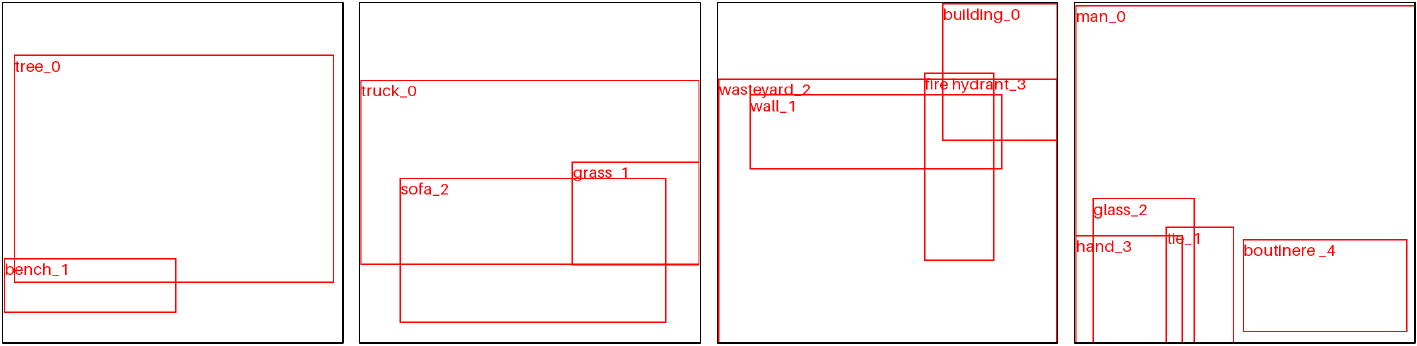}
    \caption{Examples with different number of objects in RealOcc dataset. The bounding boxes are inferred from the amodal masks in the COCOA dataset. The index after the name at the upper-left corner means the ordering from bottom to top. For example, the second example demonstrates grass occludes a truck and a sofa occludes both the track and the grass.}
    \label{fig:realocc_examples}
\end{figure*}

\noindent\textbf{Examples}.
Figure~\ref{fig:realocc_examples} shows RealOCC examples with 2 to 4 objects, covering both simple and complex occlusions.

\begin{figure*}
    \centering
    \includegraphics[width=\linewidth]{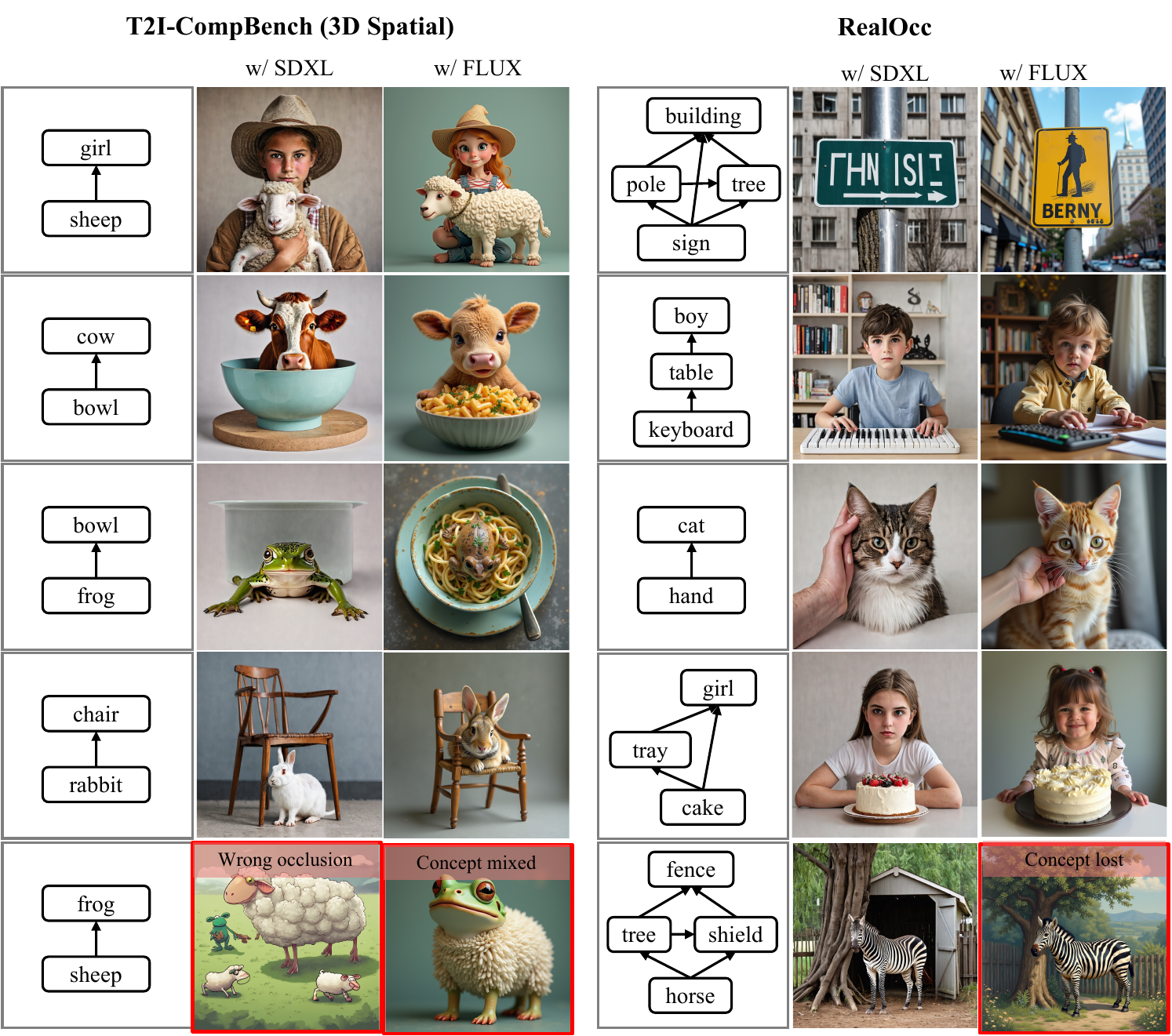}
    \caption{More results by our method, based on SDXL and FLUX. Failure cases are marked in red boxes.}
    \label{fig:more_results}
\end{figure*}

\begin{figure*}
    \centering
    \includegraphics[width=\linewidth]{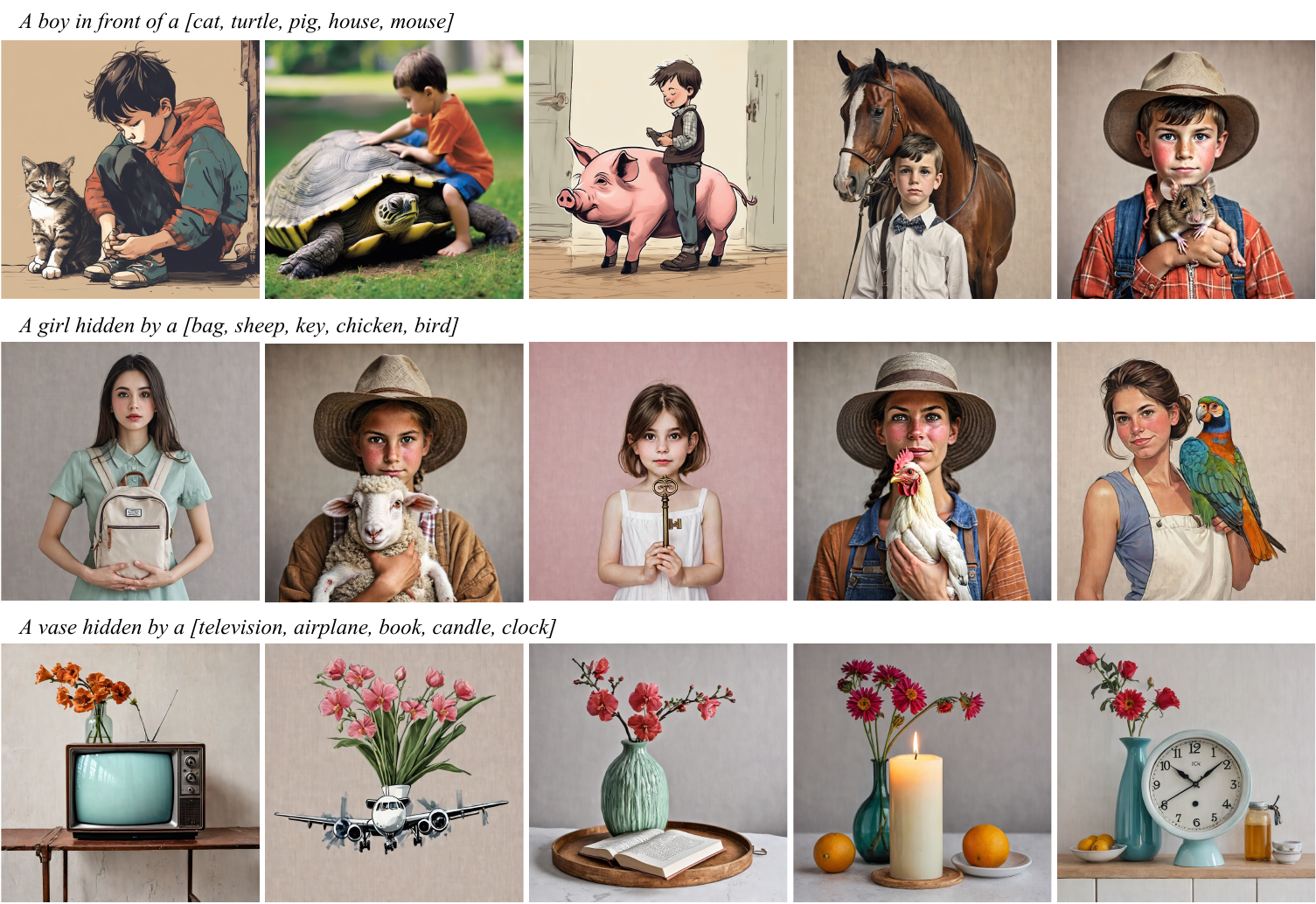}
    \caption{This figure show results of a subject occluding with other objects.}
    \label{fig:subject}
\end{figure*}

\section{More Visualizations.}
\subsection{More results}
Figure~\ref{fig:more_results} compares LaRender with SDXL and FLUX. Despite style differences, both handle occlusion well. Failure cases are also shown. Figure~\ref{fig:subject} highlights a subject occluding various concepts as an interesting example.

\subsection{Visualization of Ablation Study}
Figure~\ref{fig:more_results} illustrates the effects of different $\sigma_i(t)$ schedules and the use of cross-attention maps. A fixed opaque $\sigma_i(t)$ causes excessive opacity and object disappearance, making it only suitable in early steps. Fixing $\sigma_i(t)$ at a normal value throughout may lead to concept mixing and misaligned occlusions. In contrast, our inverse-proportional schedule generates clearer objects and occlusion relationships. Additionally, cross-attention maps help refine object contours better than bounding box masks alone.

\begin{figure*}
    \centering
    \includegraphics[width=\linewidth]{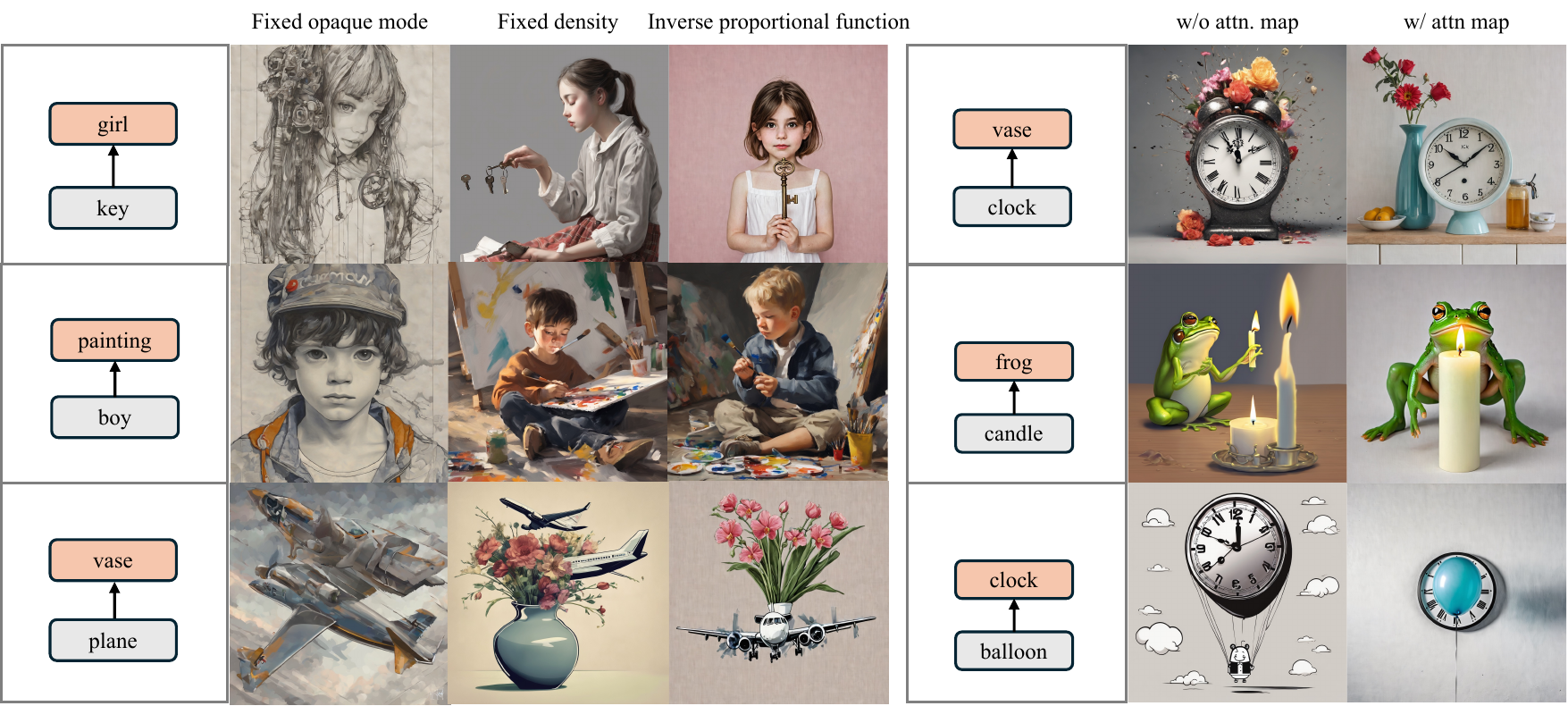}
    \caption{The visual results of ablation study, including the impact of different density schedules (left) and the presence of cross-attention maps in computing transmittance maps (right).}
    \label{fig:more_results}
\end{figure*}

\subsection{Visualization of interactions.}
Results of inter-object interactions is shown in Figure~\ref{fig:interactions}.

\subsection{Visualization of 2-element effects.}
\begin{figure*}
  \centering
  \includegraphics[width=\linewidth]{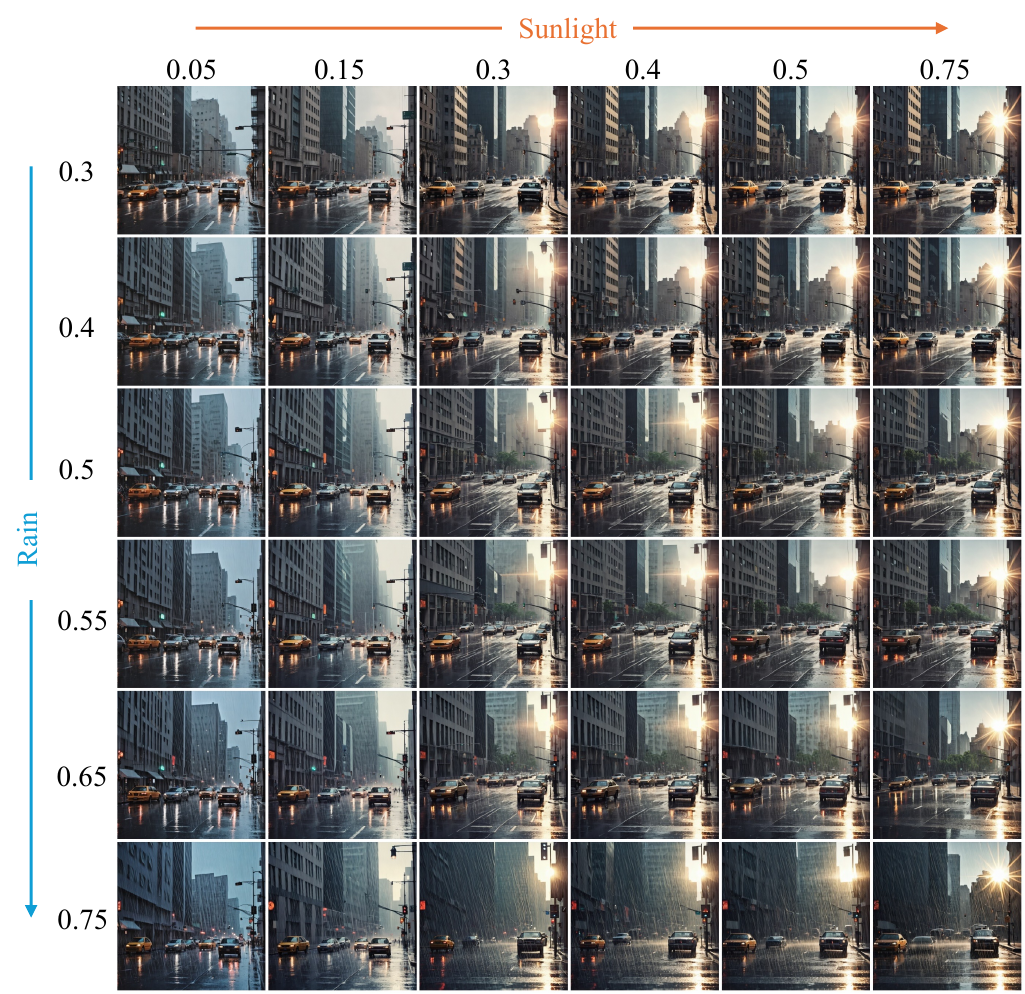}
   \caption{Grid figure of dual element changes. Please zoom in to see the changes.}
   \label{fig:grid}
\end{figure*}

To showcase LaRender’s semantic opacity control, we visualize two elements in Figure~\ref{fig:grid}, showing it can independently control multiple effects and generate high-quality results.

\vspace{20pt}

\section{LLM instructions}
We design the prompt templates in Figure~\ref{fig:subject} with instructions, in-context examples, and a test case from the user's input. The LLM follows the instruction to identify occlusion pairs and generate object layouts based on occlusion order.
\begin{figure*}
    \centering
    \includegraphics[width=\linewidth]{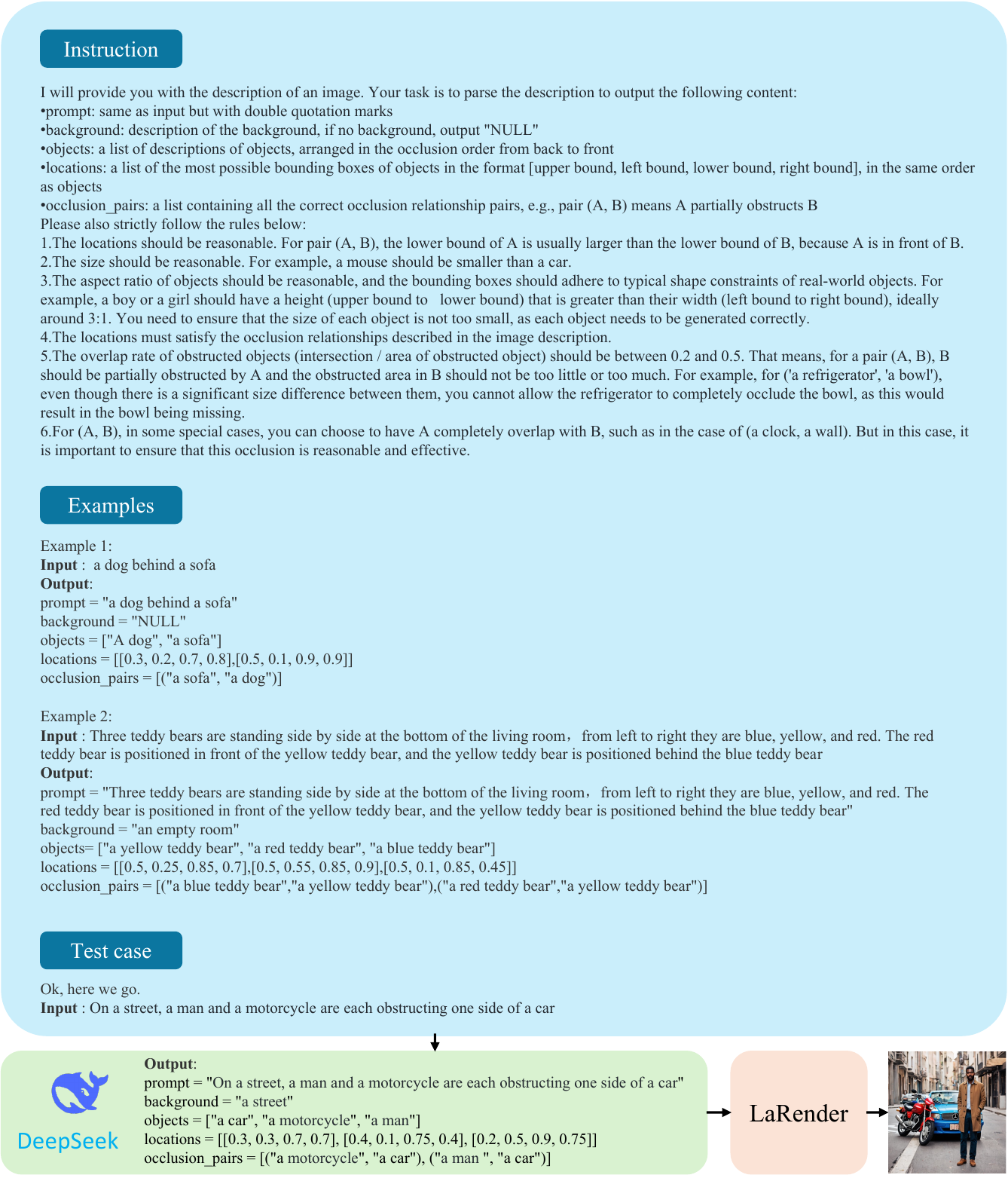}
    \caption{The instruction for DeepSeek to parse the occlusion graph.}
    \label{fig:subject}
\end{figure*}

\end{document}